\documentclass[10pt, a4paper]{article}

\usepackage{lrec-coling2024} 
\usepackage{microtype}
\usepackage{multirow}
\usepackage{graphicx}
\usepackage{bm}
\usepackage{amsmath}
\usepackage{amssymb}

\usepackage{placeins}
\usepackage{multicol}
\usepackage{caption}
\usepackage{subcaption}

\usepackage{xcolor}
\usepackage{hyperref}

\title{Enhanced Facet Generation with LLM Editing}

\name{Joosung Lee, Jinhong Kim} 

\address{NAVER\\
         \{rung.joo, kim.jinhong\}@navercorp.com\\}

\abstract{
In information retrieval, facet identification of a user query is an important task.
If a search service can recognize the facets of a user's query, it has the potential to offer users a much broader range of search results. 
Previous studies can enhance facet prediction by leveraging retrieved documents and related queries obtained through a search engine. 
However, there are challenges in extending it to other applications when a search engine operates as part of the model. 
First, search engines are constantly updated. Therefore, additional information may change during training and test, which may reduce performance. 
The second challenge is that public search engines cannot search for internal documents. 
Therefore, a separate search system needs to be built to incorporate documents from private domains within the company.
We propose two strategies that focus on a framework that can predict facets by taking only queries as input without a search engine.
The first strategy is multi-task learning to predict SERP. 
By leveraging SERP as a target instead of a source, the proposed model deeply understands queries without relying on external modules.
The second strategy is to enhance the facets by combining Large Language Model (LLM) and the small model.
Overall performance improves when small model and LLM are combined rather than facet generation individually.
 \\ \newline  }

\begin{document}

\maketitleabstract

\section{Introduction}

Search clarification has been an area of interest in information retrieval for a long time~\citep{10.1145/3020165.3020183, 10.1145/3027063.3053175}.
Users send queries with various sub-intents to the search system and expect various search results. 
These sub-intents are called facets.
For instance, facets could include "warcraft game", "warcraft movie", "warcraft book", "warcraft history" and more when a user searches for "warcraft".
Previous studies~\citep{previous1, previous2} introduce facet generation task as generating facets from the query.
If a search system can predict the query facets in advance, it can provide more diverse and higher-quality search results. 

The previous studies~\citep{previous1, previous2, previous4, previous3, previous5} demonstrated that models can improve their performance in generating various query facets by leveraging Search Engine Result Page (SERP). 
The most commonly used information in SERP is the snippet of the retrieved document. 
Constructing input with both query and document snippets provides the model with richer information, leading to improved performance in facet prediction.
However, there are several challenges to commercialize these methods. 
First, public search engines like Bing or Google are continuously updated. 
Search algorithms change over time, and user documents are continually updated. 
The external researchers cannot grasp the principles and changes of the private search algorithm.
Therefore, if the search engine is used as part of the model, SERP changes between training and testing, leading to a drop in performance.
The second challenge is that public search engines only search for public documents. 
If the systems need to create a query facet for an in-house service, the facet distribution to target will be different.
However, there is a significant cost involved in constructing a separate search engine to leverage in-house documents.
Finally, external communication is essential for SERP.
Therefore, the previous methods are difficult for customers who want on-premise services to consider.

We focus on a framework that operates independently of the search engine by using only a query as input during testing.
We propose two strategies to predict query facets without SERP.
The first is multi-task learning, which uses SERP only in the training and not in the test.
The approach of simply concatenating documents as input for training is not efficient during the test (refer to Section~\ref{sec:method}).
Therefore, we consider SERP as the target to improve the performance of our model.
The second is editing facets using LLM.
Recently, LLM has made remarkable progress since InstructGPT~\citep{ouyang2022training}, achieving high performance in a variety of tasks.
However, simply instructing an LLM to generate query facets can result in inaccurate facet generation. 
Because LLM does not know the distribution of the dataset, it is difficult to predict the facet that fits the target.
We improve performance by editing the facets predicted by the fine-tuned small model with LLM.
This is the effect of allowing LLM to generate accurate facets by informing LLM of the distribution of the dataset through a small model learned with the training dataset.
In other words, LLM editing is more effective than end-to-end generation because a fine-tuned small model generates intermediate results to the target facet.
We also demonstrate that LLM editing works effectively on the previous models as well.

\section{Related Work}
\label{sec:previous}
Previous studies leverage SERP to enhance facet generation performance.
~\citealp{previous1} proposes the \textit{NMIR} framework, which learns multiple intent representations by taking query and retrieved documents as input.
~\citealp{previous2} introduces five methods based on query and retrieved documents.
\textit{FG} (Facet Generation) generates a facet by inputting the retrieved document.
\textit{SL} (Sequence Labeling) determines whether a token is a facet through sequence labeling in the retrieved document.
\textit{EFC} (Extreme Facet Classification) considers terms that frequently appear in facets as classes and trains a classifier to find facet terms in documents.
~\citealp{previous4} proposes a permutation-invariant approach based on \textit{NMIR}.
~\citealp{previous3} proposes a method that leverages related queries in addition to documents to enhance the performance.
~\citealp{previous5} finds better facets for queries by combining external structured information in addition to documents.
\textit{SR} (Structured Relation) uses hypernyms through external knowledge (Concept Graph~\cite{ConceptGraph} and WebIsA~\cite{WebIsA}) and list structure through HTML as input.

Search Clarification is closely related to interactive search systems~\citep{related1, related2}.
This is because the search system can clarify the user's intent and provide more accurate services.
Therefore, for ambiguous queries, the search system asks a clarifying question~\citep{related3}.
~\citealp{related4} identifies the taxonomy of clarification and generates clarifying questions.
~\citealp{related5} builds a neural network model for the task of ranking clarification questions. 

Facet identification of a query is also related to learning the query representation or expansion.
Traditionally, query representations were constructed from term frequencies in search logs~\cite{salton1975vector}.
~\citealp{rocchio1971relevance, lavrenko2017relevance} introduce query representation using query expansion and relevance feedback.
~\citealp{mikolov2013distributed} learns the relationships between adjacent words in the corpus and represents words as embeddings.
Words and queries can be expressed through pre-trained language models learned with large corpora such as BERT~\cite{devlin2018bert} and GPT2~\cite{radford2019language}.
Recently, \citealp{wang2023query2doc, jagerman2023query} introduce query expansion through LLM.

\section{Task Definition}
This paper focuses on generating facets based on only queries. 
In the training, $T_{train} = \{(q_1, D_1, R_1, F_1), ..., (q_N, D_N, R_N, F_N)\}$ where $q_i$ represents a query, $D_i=\{d_{i1}, ..., d_{im}\}$ consists of snippets from $m$ retrieved documents, $R_i=\{r_{i1}, ..., r_{it}\}$ contains $t$ related queries, obtained from query logs, and $F_i=\{f_{i1}, ..., f_{ik}\}$ represents $k$ target facets.
In the test, $T_{test} = \{(q_1, F_1), ...(q_{M}, F_{M})\}$ where we generate $F_i$ from $q_i$. In contrast to the training, $D$ and $R$ cannot be used during the test.



\begin{table*}[!htb]
\centering
\resizebox{1.5\columnwidth}{!}{
\begin{tabular}{c|c|c|c|c}
\hline
Training / Test  & Term Overlap (F1) & Exact Match (F1) & Set BLEU-mean & Set BERTScore (F1) \\ \hline\hline
  QD / QD  & 0.2914                & 0.0732              & 0.3265    & 0.8794         \\
  QD / Q    & 0.1299                & 0.004               & 0.2367    & 0.8472         \\ \hline
\end{tabular}
}
\caption{Performance changes according to the input of training and test. Q means the query, D means the snippet of the document. D and Q are concatenated to form the input.}
\label{Tab:training_testing_input}
\end{table*}

\section{Method}
\label{sec:method}
The previous methods are frameworks that learn a model by using the SERP for the query as input.
Similar to the previous method, the model in Table~\ref{Tab:training_testing_input} is fine-tuned to generate facets based on BART-base~\citep{bart} by receiving queries and documents.
The performance of the fine-tuned model changes depending on the input configuration of training and testing.
The performance will significantly decrease if SERP used for training is not used in testing.
As a result, the most ideal scenario is a situation where the input configuration in training and test is the same.

Our method assumes a scenario where SERP is not available for test.
Therefore, we utilize multi-task learning by placing information in the target rather than the input of the model.
Additionally, we leverage vast knowledge by combining LLM and small models.
The term "small model" denotes a model that can be trained on a single GPU. 
In our experiments, this corresponds to BART-base. 
The term "LLM" denotes a model of size 7B or larger and can be used in various tasks through pretraining and instruction-tuning.

\subsection{Multi-task Learning}
The input is constructed by prepending special tokens to the query, which allows us to control the target to be generated.
Special tokens include [facet], [document], and [related], and are used to generate facets, snippets of documents, and related queries, respectively.
The input is as follows:

\begin{equation}
i_s = \text{concat} ([s],  \text{query})
\label{eq:input_format}
\end{equation}
where $s \in \{facet, document, related\}$.
The target output is composed of each sentence separated by "," as follows:
\begin{equation}
o_s = `` s_1, s_2, ... "
\label{eq:output_format}
\end{equation}
where $s_i$ is the target sentence corresponding to $s$. 
The loss is calculated as cross entropy as follows:
\begin{equation}
L_s = \frac{1}{N} \sum^{N}_{i=1} CE(f(i_s), o_s)
\label{eq:loss}
\end{equation}
The sum of losses based on the targets used in multi-task learning is the final loss. The trained model can generate not only facets but also documents or related queries.
The model's additional capabilities assist in generating accurate facets from queries.

\subsection{LLM Editing}
Multi-task learning improves the performance of small models without relying on the search engine. 
However, the fine-tuned model still has a limitation in not being able to leverage rich external information during the test.
Therefore, we propose a strategy to mitigate this drawback by leveraging an LLM with extensive knowledge from a large corpus.

LLM editing is a technique that refines facets generated by a fine-tuned small model. 
When LLM is instructed to generate target facets corresponding to a query, it relies on general generation capabilities.
This generative ability comes from a massive pretraining corpus and instruction tuning.
Therefore, it is difficult to convey the distribution of the desired target facets in the dataset to LLM simply through few-shot demonstrations of query and facet pairs.
On the other hand, the fine-tuned small model knows the distribution of facets to be generated because it has been learned from the training dataset.
Therefore, we provide the facets predicted by the small model to LLM to regenerate the improved facets.
With the assistance of a small model, LLM can perform modified facet identification from a state close to the target facets, making the task easier.
In other words, it is a method of leveraging the distribution of the training dataset, which is the knowledge of a fine-tuned small model.

\begin{table}[!t]
\large
\centering
\resizebox{1.0\columnwidth}{!}{
\begin{tabular}{l}
\#\#\# User: \\ The predicted facets for `\{example query1\}' are `\{predicted facets1\}'. 
\\ But the correct facets are `\{label facets1\}'.
\\ The predicted facets for `\{example query2\}' are `\{predicted facets2\}'. 
\\ But the correct facets are `\{label facets2\}'.
\\ \\ As in the example above, modify the predicted facets.
\\ \\ The predicted facets for `\{input query\}' are `\{predicted facets\}'.
\\ What are the correct facets?
\\ \\ \#\#\# Assistant:
\\ The correct facets for `\{input query\}' are
\end{tabular}
}
\caption{Prompt given to LLM. \{predicted facets\} are the output of the small model.}
\label{Tab:LLM_prompt}
\end{table}

\paragraph{Editing Prompt.} Table~\ref{Tab:LLM_prompt} shows the prompt for LLM editing the results of the small model.
We inform LLM of two-shot demonstrations (predicted facets => label facets).
If LLM does not combine small models, we instruct LLM to generate facets via few-shot or zero-shot.
In \textit{E(zero)}, LLM is instructed to generate facets without information about the dataset distribution.
In \textit{E(few)}, LLM can obtain limited information via standard prompting through few-shot demonstrations.
In Appendix~\ref{app:LLM_prompt}, prompt configuration is introduced in more detail.

\begin{table*}[!htb]
\centering
\resizebox{1.45\columnwidth}{!}{
\begin{tabular}{c|c|c|c|c}
\hline
Model   & Term Overlap (F1) & Exact Match (F1) & Set BLEU-mean & Set BERTScore (F1) \\ \hline\hline
F       & 0.2374                & 0.0284              & 0.2898     & 0.8657  \\ \hline
E(zero)      & 0.0817                & 0.0134              & 0.1512     & 0.8586  \\ 
E(few)       & 0.2101                & 0.0424              & 0.3511     & 0.8803  \\ \hline
FR+M    & 0.2519                 & 0.0337             & 0.2963    & 0.8687   \\
FR+M+E  & 0.2385                & 0.0509              & 0.3766    & 0.8807   \\ \hline
FD+M    & 0.2508                & 0.0338              & 0.2992    & 0.871    \\
FD+M+E  & 0.2381                & 0.0518              & 0.3772    & 0.8812   \\ \hline
\end{tabular}
}
\caption{Our model's performance to multi-task learning and LLM editing. F, D, and R indicate that the model was trained to generate facet, document, and related queries, respectively. +M represents multi-task learning and +E represents LLM editing combined into the small model.}
\label{Tab:multitask_learning}
\end{table*}

\section{Experiments}
We followed previous studies~\citep{previous1, previous2, previous3, previous4, previous5} and used BART-base as a small model.
ChatGPT~\citep{chatgpt} or GPT4 are private LLMs and have cost issues.
Additionally, OpenAI's models are constantly updated, making it difficult to reproduce our results, so we use open-source LLMs with public parameters.
At the time of our experiments, we used UP 30B~\citep{upstage}, which ranks high on the LLM leaderboard~\cite{open-llm-leaderboard}.


\subsection{Dataset}
MIMICS dataset~\citeplanguageresource{MIMICS}~\footnote{https://github.com/microsoft/MIMICS} is widely used in search clarification or facet generation.
MIMICS is collected from the Bing search engine and consists of three subsets.
Following previous studies, MIMICS-Click is used as a training dataset and MIMICS-Manual is used as a test dataset.
SERP was used as public data~\footnote{http://ciir.cs.umass.edu/downloads/mimics-serp/MIMICS-BingAPI-results.zip}.

\subsection{Evaluation Metric}

\subsubsection{Automatic Evaluation}

We follow the automatic metric proposed in~\citealp{previous1}.
Term Overlap indicates that the terms of generated facets and ground truth facets overlap.
Exact Match indicates whether the generated facets are identical to the ground truth facets.
Set BLEU-mean represents the average of the 1-gram, 2-gram, 3-gram, and 4-gram scores of each facet sentence.
Set BERTScore~\citep{bert-score} calculates the similarity of each facet sentence using RoBERTa-large~\citep{roberta}.
For intuitive analysis, we use a single score for metrics such as each F1 or average score.
Evaluation scripts are provided in ~\citealp{previous2}.

\begin{table}[!t]
\large
\centering
\resizebox{1.0\columnwidth}{!}{
\begin{tabular}{l}
Facets refer to the sub-intents desired by the user who 
\\ searched the query.
\\ The following are facets about "\{query\}".
\\ Which facets set is better? (without explanation)
\\ A: \{predicted facets by model A\}
\\ B: \{predicted facets by model B\}
\end{tabular}
}
\caption{Assessment prompt instructed to LLMs}
\label{Tab:LLM_eval_prompt}
\end{table}

\subsubsection{LLM-based Evaluation}
It is difficult to select the best model only by automatic evaluation.
Since automatic evaluation has four metrics, a good model depends on the metric.
Previous studies~\cite{chiang-lee-2023-large, liu-etal-2023-g} introduce that the LLM evaluator works as a good evaluator in various NLG tasks. 
LLM evaluators show a high correlation with human evaluators and show more reliable results than traditional metrics (e.g. BLEU, ROUGE, METEOR). 
Additionally, LLM assessments are highly reproducible and unaffected by previous test samples. 
LLM evaluator can be utilized in various ways, such as through a win-lose method or by computing scores. 
~\citealp{kocmi-federmann-2023-large} introduces evaluating the LLM evaluator using a scoring method in translation tasks, but it has the disadvantage that the score distribution is biased to one side.
Since we only need to determine the superiority between two compared models, we utilize the LLM evaluator in a win-lose method.
Inspired by these results, we attempt to evaluate using gemini-pro~\citep{team2023gemini} and GPT4~\citep{openai2023gpt4}, which are known to have the best performance as LLM.

Table~\ref{Tab:LLM_eval_prompt} shows the model assessment prompt.
We provided the predicted facets of models A and B to the LLM and asked which one was better. 
Therefore, the model responds with either A or B. 
However, LLM often has a different response format because it is a generative model.
To minimize the risk of not being able to verify the correct answer due to such parsing, we set it to temperature=0.1, top\_p=1.
Nonetheless, samples responding in a different format are excluded from the evaluation.

\subsection{Result and Discussion}
Table~\ref{Tab:multitask_learning} shows the experimental results of our strategies.
\textit{F} model is fine-tuned only for facet generation.
\textit{E} models are the result of instructing LLM to generate facets by providing a query and few(two)- or zero-shot demonstrations without a small model.
\textit{+M} is multi-task learning, where the model is trained to generate related queries (\textit{R}) or snippets of documents (\textit{D}) in addition to facet generation.
\textit{+E} indicates that LLM editing was performed on the results of the small model. 
Some examples and statistics of the generated facets are introduced in Appendix~\ref{app:static_example}.

\begin{table*}[!htb]
\centering
\resizebox{1.6\columnwidth}{!}{
\begin{tabular}{c|c|c|c|c|c}
\hline
Model          & Training / Test                          & Term Overlap (F1) & Exact Match (F1) & Set BLEU-mean   & Set BERTScore (F1) \\ \hline\hline
FG  & Q / Q      & 0.0664                & 0.0306              & 0.0751 & 0.8446         \\ \hline\hline
  FD+M (ours) & \multirow{2}{*}{QD / Q}      & \textbf{0.2508}                & 0.0338              & 0.2992 & 0.871         \\ 
FD+M+E (ours) &                       & 0.2381                & \textbf{0.0518}              & \textbf{0.3772} & \textbf{0.8812}         \\ \hline\hline
FG & \multirow{6}{*}{QD / QD} & \textbf{0.2919}                & \textbf{0.0707}              & 0.3544  & 0.8785         \\ 
FG+E &             & 0.2702                & 0.0694              & \textbf{0.4092} & \textbf{0.8844}         \\ 
SL &             & 0.1914                & 0.0515              & 0.2483 & 0.8748         \\ 
SL+E &             & 0.1895                & 0.0542              & 0.2618 & 0.8769         \\ 
EFC &             & 0.0515                & 0.0289              & 0.0544 & 0.423\\ 
EFC+E &             & 0.0678                & 0.028              & 0.0863 & 0.8626\\ 
\hline\hline
SR & \multirow{2}{*}{QDS / QDS} & \textbf{0.274}                & \textbf{0.0888}              & 0.4297 & \textbf{0.8903} \\ 
SR+E &                       & 0.2626                & 0.849              & \textbf{0.4302} & 0.8896  \\
\hline
\end{tabular}
}
\caption{Performance comparison between our model and other models. S in QDS indicates structured information. Bold indicates the highest performance in each test type.}
\label{Tab:total}
\end{table*}

Multi-task learning improves performance even when there is only a query in the test. 
Both \textit{FR+M} and \textit{FD+M} outperform \textit{F} in all four metrics. 
The enhanced capability of the small model to infer not only the facets but also retrieved documents or related queries leads to a better understanding of the query.
When considering all automatic metrics, \textit{FD+M} is slightly superior to \textit{FR+M}. 
As a result, we confirmed that a snippet of the document is more effective for small models than a related query.

LLM editing enhances the facets generated by the small model. 
\textit{FD+M+E} slightly reduces performance in Term Overlap compared to \textit{FD+M}, but improves performance in the other three metrics.
Overall, \textit{FD+M+E} is better than \textit{FD+M}.
\textit{FD+M+E} outperforms \textit{E(few)} in all aspects, which proves that the facets generated by the small model contribute to the target facet distribution.
\textit{E(few)} receives a distribution of the dataset via in-context learning through few-shot demonstrations, but it is very limited information.
\textit{E(zero)} has lower performance because it does not know the distribution of the dataset at all.
It is important to note that facets generated by LLM without prior information are difficult to match the target distribution.
In other words, the method of combining the small model and LLM proves to be more effective than simply fine-tuning and standard prompting. 
In the Appendix~\ref{app:other_LLM}, we show that LLM editing is effective regardless of LLM size.

\subsubsection{Comparison with Previous Methods}
Table~\ref{Tab:total} shows a comparison of the performance of our model with the previous methods.
The descriptions of comparative models are in Section~\ref{sec:previous}.
\textit{FD+M+E} demonstrates the second-best performance in Set BLEU-mean and Set BERTScore among previous models that did not combine LLM editing.
This means that \textit{FD+M+E} generates more semantically sufficient facets than \textit{FG}, which is the best in QD test type without SERP.
\textit{SR} improves performance with structured information (hypernyms and HTML) in addition to document snippets but is more dependent on SERP.
From the results of SR, we expect that leveraging structured information in multi-task learning will lead to improved performance in the future.

\subsubsection{Result of LLM-based Evaluation}
\label{sec:LLM_eval}
Table~\ref{Tab:LLM_evaluation} shows comparison results between \textit{FD+M+E} and other models.
We selected \textit{FD+M}, \textit{FG(QD/QD)} and \textit{SR} as comparison models.
The numbers in the cell are the percentages that \textit{FD+M+E} won in competition with other models.
For example, GPT4 determines that \textit{FD+M+E} is better than \textit{SR} for 63.86\% of the test data.
Both LLMs determine that \textit{FD+M+E} performs better than the other three models.
These results prove that \textit{FD+M+E} is a more effective method than previous methods using SERP.
In other words, \textit{FD+M+E} is better than the previous SoTA in terms of LLM-based evaluation perspective.
In particular, the higher win rate compared to \textit{FD+M} indicates that LLM editing is an important factor.
Since LLM evaluators are known to be more relevant to human evaluators than traditional metrics, our method is considered to have achieved state-of-the-art performance without leveraging SERP.

\subsubsection{Combine LLM Editing with Previous Methods}
We applied LLM editing to previous models that used SERP as input.
Table~\ref{Tab:total} shows that the effect of LLM editing varies depending on the performance of the models.
Similar to Table~\ref{Tab:multitask_learning}, LLM editing tends to improve overall Set BLUE-mean and Set BERTScore performance.
However, LLM editing tends to decrease Term Overlap performance, and changes in Exact Match performance vary depending on the model.
LLM editing substantially enhances the performance of models such as \textit{EFC}, which demonstrate inferior performance overall.
We demonstrate that LLM editing is an effective technique for regenerating semantic facets regardless of the small model.


\begin{table}[!t]
\centering
\resizebox{0.65\columnwidth}{!}{
\begin{tabular}{ccc}
\hline
\multicolumn{3}{c}{FD+M+E vs Comparison model}                                               \\ \hline
\multicolumn{1}{c|}{Comparison model}              & \multicolumn{1}{c|}{Gemini-pro} & GPT4  \\ \hline\hline
\multicolumn{1}{c|}{FD+M}  & \multicolumn{1}{c|}{72.33}      & 90.65 \\ \hline
\multicolumn{1}{c|}{FG (QD/QD)} & \multicolumn{1}{c|}{59.05}      & 78.59 \\ \hline
\multicolumn{1}{c|}{SR}               & \multicolumn{1}{c|}{74.08}      & 63.86 \\ \hline
\end{tabular}
}
\caption{Win ratio of FD+M+E that competed with the other models. Gemini-pro and GPT4 are used as LLM evaluators.}
\label{Tab:LLM_evaluation}
\end{table}

\section{Conclusion}
The proposed method generates facets using only queries, which eliminates the dependency on search engines.
To address the limitation of not being able to utilize SERP as input, we propose two strategies: multi-task learning and LLM editing.
Multi-task learning helps the small model better understand the query.
LLM receives prior information from a small model and generates improved facets.
Even without SERP, \textit{FD+M+E} shows similar performance to \textit{FG} in automatic evaluation and achieves the best performance in LLM-based evaluation.
LLM editing is a way to effectively combine small models and LLMs in various NLP tasks, rather than using them separately.
Therefore, our method can be extended to various NLP tasks.


\nocite{*}
\section{Bibliographical References}\label{sec:reference}

\bibliographystyle{lrec-coling2024-natbib}
\bibliography{lrec-coling2024-example}

\begin{thebibliography}{1}
\expandafter\ifx\csname natexlab\endcsname\relax\def\natexlab#1{#1}\fi

\bibitem[{Zamani et~al.(2020)Zamani, Lueck, Chen, Quispe, Luu, and
  Craswell}]{MIMICS}
Hamed Zamani, Gord Lueck, Everest Chen, Rodolfo Quispe, Flint Luu, and Nick
  Craswell. 2020.
\newblock \href {https://doi.org/10.1145/3340531.3412772} {Mimics: A
  large-scale data collection for search clarification}.
\newblock In \emph{Proceedings of the 29th ACM International Conference on
  Information \& Knowledge Management}, CIKM '20, page 3189–3196, New York,
  NY, USA. Association for Computing Machinery.

\end{thebibliography}


\begin{thebibliography}{40}
\expandafter\ifx\csname natexlab\endcsname\relax\def\natexlab#1{#1}\fi

\bibitem[{Aliannejadi et~al.(2021)Aliannejadi, Kiseleva, Chuklin, Dalton, and
  Burtsev}]{related2}
Mohammad Aliannejadi, Julia Kiseleva, Aleksandr Chuklin, Jeff Dalton, and
  Mikhail Burtsev. 2021.
\newblock \href {https://doi.org/10.18653/v1/2021.emnlp-main.367} {Building and
  evaluating open-domain dialogue corpora with clarifying questions}.
\newblock In \emph{Proceedings of the 2021 Conference on Empirical Methods in
  Natural Language Processing}, pages 4473--4484, Online and Punta Cana,
  Dominican Republic. Association for Computational Linguistics.

\bibitem[{Aliannejadi et~al.(2019)Aliannejadi, Zamani, Crestani, and
  Croft}]{related3}
Mohammad Aliannejadi, Hamed Zamani, Fabio Crestani, and W.~Bruce Croft. 2019.
\newblock \href {https://doi.org/10.1145/3331184.3331265} {Asking clarifying
  questions in open-domain information-seeking conversations}.
\newblock In \emph{Proceedings of the 42nd International ACM SIGIR Conference
  on Research and Development in Information Retrieval}, SIGIR'19, page
  475–484, New York, NY, USA. Association for Computing Machinery.

\bibitem[{Beeching et~al.(2023)Beeching, Fourrier, Habib, Han, Lambert, Rajani,
  Sanseviero, Tunstall, and Wolf}]{open-llm-leaderboard}
Edward Beeching, Clémentine Fourrier, Nathan Habib, Sheon Han, Nathan Lambert,
  Nazneen Rajani, Omar Sanseviero, Lewis Tunstall, and Thomas Wolf. 2023.
\newblock Open llm leaderboard.
\newblock
  \url{https://huggingface.co/spaces/HuggingFaceH4/open_llm_leaderboard}.

\bibitem[{Chia et~al.(2023)Chia, Chen, Tuan, Poria, and
  Bing}]{chia2023contrastive}
Yew~Ken Chia, Guizhen Chen, Luu~Anh Tuan, Soujanya Poria, and Lidong Bing.
  2023.
\newblock Contrastive chain-of-thought prompting.
\newblock \emph{arXiv preprint arXiv:2311.09277}.

\bibitem[{Chiang and Lee(2023)}]{chiang-lee-2023-large}
Cheng-Han Chiang and Hung-yi Lee. 2023.
\newblock \href {https://doi.org/10.18653/v1/2023.acl-long.870} {Can large
  language models be an alternative to human evaluations?}
\newblock In \emph{Proceedings of the 61st Annual Meeting of the Association
  for Computational Linguistics (Volume 1: Long Papers)}, pages 15607--15631,
  Toronto, Canada. Association for Computational Linguistics.

\bibitem[{Devlin et~al.(2018)Devlin, Chang, Lee, and
  Toutanova}]{devlin2018bert}
Jacob Devlin, Ming-Wei Chang, Kenton Lee, and Kristina Toutanova. 2018.
\newblock Bert: Pre-training of deep bidirectional transformers for language
  understanding.
\newblock \emph{arXiv preprint arXiv:1810.04805}.

\bibitem[{Hashemi et~al.(2021)Hashemi, Zamani, and Croft}]{previous1}
Helia Hashemi, Hamed Zamani, and W.~Bruce Croft. 2021.
\newblock \href {https://doi.org/10.1145/3459637.3482445} {Learning multiple
  intent representations for search queries}.
\newblock In \emph{Proceedings of the 30th ACM International Conference on
  Information \& Knowledge Management}, CIKM '21, page 669–679, New York, NY,
  USA. Association for Computing Machinery.

\bibitem[{Hashemi et~al.(2022)Hashemi, Zamani, and Croft}]{previous4}
Helia Hashemi, Hamed Zamani, and W.~Bruce Croft. 2022.
\newblock \href {https://doi.org/10.1145/3511808.3557666} {Stochastic
  optimization of text set generation for learning multiple query intent
  representations}.
\newblock In \emph{Proceedings of the 31st ACM International Conference on
  Information \& Knowledge Management}, CIKM '22, page 4003–4008, New York,
  NY, USA. Association for Computing Machinery.

\bibitem[{Hsieh et~al.(2023)Hsieh, Li, Yeh, Nakhost, Fujii, Ratner, Krishna,
  Lee, and Pfister}]{distilling}
Cheng-Yu Hsieh, Chun-Liang Li, Chih-kuan Yeh, Hootan Nakhost, Yasuhisa Fujii,
  Alex Ratner, Ranjay Krishna, Chen-Yu Lee, and Tomas Pfister. 2023.
\newblock \href {https://doi.org/10.18653/v1/2023.findings-acl.507} {Distilling
  step-by-step! outperforming larger language models with less training data
  and smaller model sizes}.
\newblock In \emph{Findings of the Association for Computational Linguistics:
  ACL 2023}, pages 8003--8017, Toronto, Canada. Association for Computational
  Linguistics.

\bibitem[{HyperbeeAI(2023)}]{hyperbee}
HyperbeeAI. 2023.
\newblock Hyperbeeai/tulpar-7b-v0.
\newblock \url{https://huggingface.co/HyperbeeAI/Tulpar-7b-v0}.

\bibitem[{Jagerman et~al.(2023)Jagerman, Zhuang, Qin, Wang, and
  Bendersky}]{jagerman2023query}
Rolf Jagerman, Honglei Zhuang, Zhen Qin, Xuanhui Wang, and Michael Bendersky.
  2023.
\newblock Query expansion by prompting large language models.
\newblock \emph{arXiv preprint arXiv:2305.03653}.

\bibitem[{Kocmi and Federmann(2023)}]{kocmi-federmann-2023-large}
Tom Kocmi and Christian Federmann. 2023.
\newblock \href {https://aclanthology.org/2023.eamt-1.19} {Large language
  models are state-of-the-art evaluators of translation quality}.
\newblock In \emph{Proceedings of the 24th Annual Conference of the European
  Association for Machine Translation}, pages 193--203, Tampere, Finland.
  European Association for Machine Translation.

\bibitem[{Lavrenko and Croft(2017)}]{lavrenko2017relevance}
Victor Lavrenko and W~Bruce Croft. 2017.
\newblock Relevance-based language models.
\newblock In \emph{ACM SIGIR Forum}, volume~51, pages 260--267. ACM New York,
  NY, USA.

\bibitem[{Lee et~al.(2023)Lee, Hunter, Ruiz, Goodson, Lian, Wang, Pentland,
  Cook, Vong, and "Teknium"}]{hunterlee2023orcaplaty1}
Ariel~N. Lee, Cole~J. Hunter, Nataniel Ruiz, Bleys Goodson, Wing Lian, Guan
  Wang, Eugene Pentland, Austin Cook, Chanvichet Vong, and "Teknium". 2023.
\newblock Openorcaplatypus: Llama2-13b model instruct-tuned on filtered
  openorcav1 gpt-4 dataset and merged with divergent stem and logic dataset
  model.
\newblock \url{https://huggingface.co/Open-Orca/OpenOrca-Platypus2-13B}.

\bibitem[{Lewis et~al.(2019)Lewis, Liu, Goyal, Ghazvininejad, Mohamed, Levy,
  Stoyanov, and Zettlemoyer}]{bart}
Mike Lewis, Yinhan Liu, Naman Goyal, Marjan Ghazvininejad, Abdelrahman Mohamed,
  Omer Levy, Veselin Stoyanov, and Luke Zettlemoyer. 2019.
\newblock \href {http://arxiv.org/abs/1910.13461} {{BART:} denoising
  sequence-to-sequence pre-training for natural language generation,
  translation, and comprehension}.
\newblock \emph{CoRR}, abs/1910.13461.

\bibitem[{Liu(2023)}]{previous3}
Xinyu Liu. 2023.
\newblock \href {https://doi.org/10.1109/ICBDA57405.2023.10104948} {Query
  sub-intent mining by incorporating search results with query logs for
  information retrieval}.
\newblock In \emph{2023 IEEE 8th International Conference on Big Data Analytics
  (ICBDA)}, pages 180--186.

\bibitem[{Liu et~al.(2023)Liu, Iter, Xu, Wang, Xu, and Zhu}]{liu-etal-2023-g}
Yang Liu, Dan Iter, Yichong Xu, Shuohang Wang, Ruochen Xu, and Chenguang Zhu.
  2023.
\newblock \href {https://doi.org/10.18653/v1/2023.emnlp-main.153} {{G}-eval:
  {NLG} evaluation using gpt-4 with better human alignment}.
\newblock In \emph{Proceedings of the 2023 Conference on Empirical Methods in
  Natural Language Processing}, pages 2511--2522, Singapore. Association for
  Computational Linguistics.

\bibitem[{Liu et~al.(2019)Liu, Ott, Goyal, Du, Joshi, Chen, Levy, Lewis,
  Zettlemoyer, and Stoyanov}]{roberta}
Yinhan Liu, Myle Ott, Naman Goyal, Jingfei Du, Mandar Joshi, Danqi Chen, Omer
  Levy, Mike Lewis, Luke Zettlemoyer, and Veselin Stoyanov. 2019.
\newblock \href {http://arxiv.org/abs/1907.11692} {Roberta: {A} robustly
  optimized {BERT} pretraining approach}.
\newblock \emph{CoRR}, abs/1907.11692.

\bibitem[{Mikolov et~al.(2013)Mikolov, Sutskever, Chen, Corrado, and
  Dean}]{mikolov2013distributed}
Tomas Mikolov, Ilya Sutskever, Kai Chen, Greg~S Corrado, and Jeff Dean. 2013.
\newblock Distributed representations of words and phrases and their
  compositionality.
\newblock \emph{Advances in neural information processing systems}, 26.

\bibitem[{OpenAI(2022)}]{chatgpt}
OpenAI. 2022.
\newblock \href {https://openai.com/blog/chatgpt} {Chatgpt: Optimizing language
  models for dialogue}.

\bibitem[{OpenAI(2023)}]{openai2023gpt4}
OpenAI. 2023.
\newblock \href {http://arxiv.org/abs/2303.08774} {Gpt-4 technical report}.

\bibitem[{Ouyang et~al.(2022)Ouyang, Wu, Jiang, Almeida, Wainwright, Mishkin,
  Zhang, Agarwal, Slama, Ray et~al.}]{ouyang2022training}
Long Ouyang, Jeffrey Wu, Xu~Jiang, Diogo Almeida, Carroll Wainwright, Pamela
  Mishkin, Chong Zhang, Sandhini Agarwal, Katarina Slama, Alex Ray, et~al.
  2022.
\newblock Training language models to follow instructions with human feedback.
\newblock \emph{Advances in Neural Information Processing Systems},
  35:27730--27744.

\bibitem[{Radford et~al.(2019)Radford, Wu, Child, Luan, Amodei, Sutskever
  et~al.}]{radford2019language}
Alec Radford, Jeffrey Wu, Rewon Child, David Luan, Dario Amodei, Ilya
  Sutskever, et~al. 2019.
\newblock Language models are unsupervised multitask learners.
\newblock \emph{OpenAI blog}, 1(8):9.

\bibitem[{Radlinski and Craswell(2017)}]{10.1145/3020165.3020183}
Filip Radlinski and Nick Craswell. 2017.
\newblock \href {https://doi.org/10.1145/3020165.3020183} {A theoretical
  framework for conversational search}.
\newblock In \emph{Proceedings of the 2017 Conference on Conference Human
  Information Interaction and Retrieval}, CHIIR '17, page 117–126, New York,
  NY, USA. Association for Computing Machinery.

\bibitem[{Rao and Daum{\'e}~III(2018)}]{related5}
Sudha Rao and Hal Daum{\'e}~III. 2018.
\newblock \href {https://doi.org/10.18653/v1/P18-1255} {Learning to ask good
  questions: Ranking clarification questions using neural expected value of
  perfect information}.
\newblock In \emph{Proceedings of the 56th Annual Meeting of the Association
  for Computational Linguistics (Volume 1: Long Papers)}, pages 2737--2746,
  Melbourne, Australia. Association for Computational Linguistics.

\bibitem[{Rocchio~Jr(1971)}]{rocchio1971relevance}
Joseph~John Rocchio~Jr. 1971.
\newblock Relevance feedback in information retrieval.
\newblock \emph{The SMART retrieval system: experiments in automatic document
  processing}.

\bibitem[{Salton et~al.(1975)Salton, Wong, and Yang}]{salton1975vector}
Gerard Salton, Anita Wong, and Chung-Shu Yang. 1975.
\newblock A vector space model for automatic indexing.
\newblock \emph{Communications of the ACM}, 18(11):613--620.

\bibitem[{Samarinas et~al.(2022)Samarinas, Dharawat, and Zamani}]{previous2}
Chris Samarinas, Arkin Dharawat, and Hamed Zamani. 2022.
\newblock \href {https://doi.org/10.1145/3539813.3545138} {Revisiting open
  domain query facet extraction and generation}.
\newblock In \emph{Proceedings of the 2022 ACM SIGIR International Conference
  on Theory of Information Retrieval}, ICTIR '22, page 43–50, New York, NY,
  USA. Association for Computing Machinery.

\bibitem[{Seitner et~al.(2016)Seitner, Bizer, Eckert, Faralli, Meusel,
  Paulheim, and Ponzetto}]{WebIsA}
Julian Seitner, Christian Bizer, Kai Eckert, Stefano Faralli, Robert Meusel,
  Heiko Paulheim, and Simone~Paolo Ponzetto. 2016.
\newblock \href {https://aclanthology.org/L16-1056} {A large {D}ata{B}ase of
  hypernymy relations extracted from the web.}
\newblock In \emph{Proceedings of the Tenth International Conference on
  Language Resources and Evaluation ({LREC}'16)}, pages 360--367,
  Portoro{\v{z}}, Slovenia. European Language Resources Association (ELRA).

\bibitem[{Sekuli\'{c} et~al.(2021)Sekuli\'{c}, Aliannejadi, and
  Crestani}]{related1}
Ivan Sekuli\'{c}, Mohammad Aliannejadi, and Fabio Crestani. 2021.
\newblock \href {https://doi.org/10.1145/3471158.3472257} {Towards facet-driven
  generation of clarifying questions for conversational search}.
\newblock In \emph{Proceedings of the 2021 ACM SIGIR International Conference
  on Theory of Information Retrieval}, ICTIR '21, page 167–175, New York, NY,
  USA. Association for Computing Machinery.

\bibitem[{Team et~al.(2023)Team, Anil, Borgeaud, Wu, Alayrac, Yu, Soricut,
  Schalkwyk, Dai, Hauth et~al.}]{team2023gemini}
Gemini Team, Rohan Anil, Sebastian Borgeaud, Yonghui Wu, Jean-Baptiste Alayrac,
  Jiahui Yu, Radu Soricut, Johan Schalkwyk, Andrew~M Dai, Anja Hauth, et~al.
  2023.
\newblock Gemini: a family of highly capable multimodal models.
\newblock \emph{arXiv preprint arXiv:2312.11805}.

\bibitem[{Upstage(2023)}]{upstage}
Upstage. 2023.
\newblock upstage/llama-30b-instruct-2048.
\newblock \url{https://huggingface.co/upstage/llama-30b-instruct-2048}.

\bibitem[{Vtyurina et~al.(2017)Vtyurina, Savenkov, Agichtein, and
  Clarke}]{10.1145/3027063.3053175}
Alexandra Vtyurina, Denis Savenkov, Eugene Agichtein, and Charles L.~A. Clarke.
  2017.
\newblock \href {https://doi.org/10.1145/3027063.3053175} {Exploring
  conversational search with humans, assistants, and wizards}.
\newblock In \emph{Proceedings of the 2017 CHI Conference Extended Abstracts on
  Human Factors in Computing Systems}, CHI EA '17, page 2187–2193, New York,
  NY, USA. Association for Computing Machinery.

\bibitem[{Wang et~al.(2023)Wang, Yang, and Wei}]{wang2023query2doc}
Liang Wang, Nan Yang, and Furu Wei. 2023.
\newblock \href {https://doi.org/10.18653/v1/2023.emnlp-main.585} {Query2doc:
  Query expansion with large language models}.
\newblock In \emph{Proceedings of the 2023 Conference on Empirical Methods in
  Natural Language Processing}, pages 9414--9423, Singapore. Association for
  Computational Linguistics.

\bibitem[{Wang et~al.(2015)Wang, Wang, Wen, and Xiao}]{ConceptGraph}
Zhongyuan Wang, Haixun Wang, Ji-Rong Wen, and Yanghua Xiao. 2015.
\newblock \href {https://doi.org/10.1145/2806416.2806533} {An inference
  approach to basic level of categorization}.
\newblock In \emph{Proceedings of the 24th ACM International on Conference on
  Information and Knowledge Management}, CIKM '15, page 653–662, New York,
  NY, USA. Association for Computing Machinery.

\bibitem[{Wei et~al.(2022)Wei, Wang, Schuurmans, Bosma, Xia, Chi, Le, Zhou
  et~al.}]{wei2022chain}
Jason Wei, Xuezhi Wang, Dale Schuurmans, Maarten Bosma, Fei Xia, Ed~Chi, Quoc~V
  Le, Denny Zhou, et~al. 2022.
\newblock Chain-of-thought prompting elicits reasoning in large language
  models.
\newblock \emph{Advances in Neural Information Processing Systems},
  35:24824--24837.

\bibitem[{Zamani et~al.(2020{\natexlab{a}})Zamani, Dumais, Craswell, Bennett,
  and Lueck}]{related4}
Hamed Zamani, Susan Dumais, Nick Craswell, Paul Bennett, and Gord Lueck.
  2020{\natexlab{a}}.
\newblock \href {https://doi.org/10.1145/3366423.3380126} {Generating
  clarifying questions for information retrieval}.
\newblock In \emph{Proceedings of The Web Conference 2020}, WWW '20, page
  418–428, New York, NY, USA. Association for Computing Machinery.

\bibitem[{Zamani et~al.(2020{\natexlab{b}})Zamani, Lueck, Chen, Quispe, Luu,
  and Craswell}]{MIMICS}
Hamed Zamani, Gord Lueck, Everest Chen, Rodolfo Quispe, Flint Luu, and Nick
  Craswell. 2020{\natexlab{b}}.
\newblock \href {https://doi.org/10.1145/3340531.3412772} {Mimics: A
  large-scale data collection for search clarification}.
\newblock In \emph{Proceedings of the 29th ACM International Conference on
  Information \& Knowledge Management}, CIKM '20, page 3189–3196, New York,
  NY, USA. Association for Computing Machinery.

\bibitem[{Zhang* et~al.(2020)Zhang*, Kishore*, Wu*, Weinberger, and
  Artzi}]{bert-score}
Tianyi Zhang*, Varsha Kishore*, Felix Wu*, Kilian~Q. Weinberger, and Yoav
  Artzi. 2020.
\newblock \href {https://openreview.net/forum?id=SkeHuCVFDr} {Bertscore:
  Evaluating text generation with bert}.
\newblock In \emph{International Conference on Learning Representations}.

\bibitem[{Zhao et~al.(2023)Zhao, Dou, Guo, Cao, and Cheng}]{previous5}
Ziliang Zhao, Zhicheng Dou, Yu~Guo, Zhao Cao, and Xiaohua Cheng. 2023.
\newblock \href {https://doi.org/10.1145/3580305.3599389} {Improving search
  clarification with structured information extracted from search results}.
\newblock In \emph{Proceedings of the 29th ACM SIGKDD Conference on Knowledge
  Discovery and Data Mining}, KDD '23, page 3549–3558, New York, NY, USA.
  Association for Computing Machinery.

\end{thebibliography}

\section{Language Resource References}
\label{lr:ref}

\bibliographystylelanguageresource{lrec-coling2024-natbib}
\bibliographylanguageresource{languageresource}

\appendix

\FloatBarrier
\begin{figure*}[!t]
    \centering 
    \includegraphics[width=2.0\columnwidth]{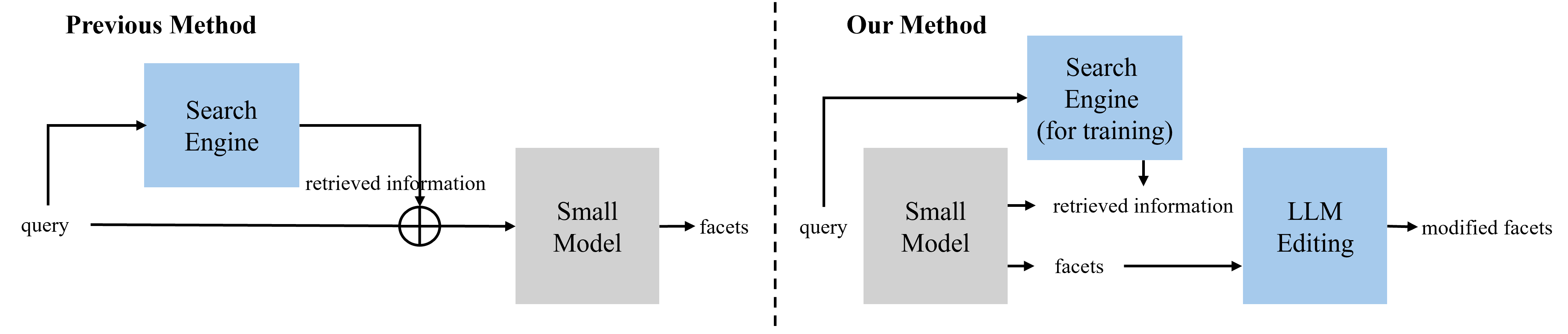}
    \caption{Overview of previous methods and our method.}
    \label{fig:model_overview}
\end{figure*}

\begin{table*}[!t]
\centering
\resizebox{2.0\columnwidth}{!}{
\begin{tabular}{c|c|c|c}
\hline
query    & FD+M                                                                                             & FD+M+E                                                                                                                  & ground-truth                                                                                          \\ \hline\hline
carrots  & carrots for sale, carrots care                                                                   & \begin{tabular}[c]{@{}c@{}}carrots nutrition, carrots health benefits, \\ carrots recipes\end{tabular}                  & \begin{tabular}[c]{@{}c@{}}grow carrots, cook carrots, \\ store carrots, freeze carrots\end{tabular} \\ \hline
orange   & \begin{tabular}[c]{@{}c@{}}orange tree, orange flower\end{tabular}               & \begin{tabular}[c]{@{}c@{}}orange fruit, orange juice, \\ orange tree, orange flower\end{tabular}                       & \begin{tabular}[c]{@{}c@{}}orange the color, orange the fruit, \\ orange the company\end{tabular}    \\ \hline
firewall & \begin{tabular}[c]{@{}c@{}}firewall windows 10, windows 7, \\ windows 8, windows xp\end{tabular} & \begin{tabular}[c]{@{}c@{}}firewall types, firewall software, \\ firewall hardware, firewall configuration\end{tabular} & firewall hardware, firewall the movie                                                                \\ \hline
\end{tabular}
}
\caption{Examples generated by FD+M and FD+M+E}
\label{Tab:modified}
\end{table*}

\begin{table*}[!t]
\centering
\resizebox{1.6\columnwidth}{!}{
\begin{tabular}{c|c|c|c|c}
\hline
Model  & \begin{tabular}[c]{@{}c@{}}average number \\of a set\end{tabular} & \begin{tabular}[c]{@{}c@{}}average length of \\ generated facets\end{tabular} & \begin{tabular}[c]{@{}c@{}}proportion of query  \\ included in facet\end{tabular}  & \begin{tabular}[c]{@{}c@{}}proportion of duplicate  \\ facets in a set\end{tabular} \\ \hline\hline
FD+M   & 2.39 & 17.9 & 61.87 & 0.038 \\ \hline
FD+M+E & 4.13 & 16.67 & 57.33 & 0.013 \\ \hline
groud-truth & 3.01 & 15 & 47.96 & 0 \\ \hline
\end{tabular}
}
\caption{Statistics of facets generated with FD+M, FD+M+E,  and ground-truth}
\label{Tab:staistic}
\end{table*}
\FloatBarrier

\begin{table}[!t]
\large
\centering
\resizebox{1.0\columnwidth}{!}{
\begin{tabular}{l}
\#\#\# User: \\ The facets for `\{query\}' are `\{facets\}'. As in the format above,
\\ generate facets related to the query within 5, separated by `,'. 
\\ \\ \#\#\# Assistant:
\\ The facets for `\{input query\}' are
\end{tabular}
}
\caption{Prompt for \textit{E(zero)} model. This prompt instructs LLM to generate the facet without any prior information, where \{query\} and \{facets\} are not examples, but strings themselves.}
\label{Tab:LLM_zeroshot_prompt}
\end{table}

\begin{table}[!t]
\large
\centering
\resizebox{1.0\columnwidth}{!}{
\begin{tabular}{l}
\#\#\# User: \\ The facets for `\{example query1\}' are `\{correct facets1\}'. 
\\ The facets for `\{example query2\}' are `\{correct facets2\}'. 
\\ \\ \#\#\# Assistant:
\\ The correct facets for `\{input query\}' are
\end{tabular}
}
\caption{Prompt for \textit{E(few)} model. It shows only few-shot demonstrations, not facet modifications.}
\label{Tab:LLM_fewshot_prompt}
\end{table}

\begin{table*}[!htb]
\centering
\resizebox{1.9\columnwidth}{!}{
\begin{tabular}{c|c|c|c|c|c}
\hline
LLM                     & Model & Term Overlap (F1) & Exact Match (F1) & Set BLEU-mean & Set BERTScore (F1) \\ \hline\hline
\multirow{2}{*}{HB 7B~\citep{hyperbee}}  & E(few)     & 0.1905            & 0.0241           & 0.206        & 0.8691             \\
                        & FD+M+E   & \textbf{0.2495}            & 0.049            & 0.3657        & 0.878             \\ 
                        \hline\hline     
\multirow{2}{*}{OO 13B~\citep{hunterlee2023orcaplaty1}} & E(few)     & 0.1852            & 0.0355            & 0.3148        & 0.8677             \\    
                        & FD+M+E   & 0.2477            & 0.0477           & 0.3721        & 0.878             \\ 
                        \hline\hline
\multirow{2}{*}{UP 30B~\citep{upstage}} & E(few)     & 0.2101            & 0.0424            & 0.3511        & 0.8803             \\
                        & FD+M+E   & 0.2381            & \textbf{0.0518}           & \textbf{0.3772}        & \textbf{0.8812}             \\ \hline
\end{tabular}
}
\caption{Performance based on other LLMs. Bold indicates the best performance.}
\label{Tab:LLM_change}
\end{table*}

\begin{table}[!htb]
\centering
\resizebox{1.0\columnwidth}{!}{
\begin{tabular}{c|c|c|c|c|c}
\hline
LLM    & Average & ARC   & HellaSwag & MMLU  & TruthfulQA \\ \hline\hline
UP 30B & 67.02   & 64.93 & 84.94     & 61.9  & 56.3       \\ \hline
OO 13B & 63.19   & 61.52 & 82.27     & 58.85 & 50.11      \\ \hline
HA 7B  & 59.89   & 56.31 & 79.01     & 52.55 & 51.68      \\ \hline
\end{tabular}
}
\caption{LLM benchmark performance reported on LLM leaderboard}
\label{Tab:LLM_performance}
\end{table}

\section{Details of Editing Prompt}
\label{app:LLM_prompt}
Table~\ref{Tab:LLM_zeroshot_prompt} and Table~\ref{Tab:LLM_fewshot_prompt} show prompts for \textit{E(zero)} and \textit{E(few)}, respectively.
Since \textit{E(zero)} lacks information about facets, it provides structured information about the format and number of facets in the "\#\#\# USER" section.
We also experiment with other LLM sizes in Appendix~\ref{app:other_LLM}.
In the prompt of OO 13B, special phrases are guided to use "Instruction" and "Response".
Therefore, the prompt of OO 13B is composed of "User" and "Assistant" replaced with "Instruction" and "Response", respectively.
We used the same generation configuration settings for all LLMs.

Chain-of-thought prompting (CoT)~\cite{wei2022chain} and contrastive chain-of-thought prompting (CCoT)~\cite{chia2023contrastive} are effective techniques in LLM prompting. CoT enhances performance by prompting LLM to generate rationales along with responses. 
Our goal is not to find a better prompt for facet identification but to improve performance by combining the fine-tuned small model and LLM.
Therefore, we leave the exploration of better prompts for future research.

\section{Statistics of Generated Facet Set}
\label{app:static_example}
Table~\ref{Tab:modified} shows examples of our model.
For carrots, LLM editing modifies facets through suffixes of nutrition, recipes, and health benefits.
For orange, LLM editing adds two facets while maintaining some of the facets predicted by \textit{FD+M}.
For firewall, LLM editing adds a firewall prefix and creates a wider range of facets.

Table~\ref{Tab:staistic} shows facet statistics of the proposed model.
LLM editing increases the number of existing facets from an average of 2.39 (\textit{FD+M}) to 4.13 (\textit{FD+M+E}) because LLM finds intentions that \textit{FD+M} could not cover.
The average length of each generated facet does not differ significantly. 
Furthermore, we also measure the extent to which the query string is important in constructing facets.
As shown in the example in Table~\ref{Tab:modified}, the intention of the facet including the query string is intuitively clear.
However, in terms of measurement results, the order of query inclusion rates is higher for \textit{FD+M > FD+M+E > ground-truth}.
In other words, we confirmed that query string is not an essential element in configuring facets, and LLM editing removes queries from facets as needed.
For example, there are cases where the ground-truth sample has the query "internet explorer" and the facet is "windows 10".
When the generated facet set contains duplicate facets, it can negatively impact the overall performance of the facet set. 
As a result of measurement, LLM editing further improves the facet set by removing duplicate facets.
Through the distribution of the generated facet set, we gain insights that limiting the number of duplicate facets and the total number of generated facets can improve future performance.

\section{Effects of LLM Size}
\label{app:other_LLM}
We conducted experiments on LLM(7B, 13B) in addition to LLM(30B).
Table~\ref{Tab:LLM_performance} shows LLM benchmark performance.
As the size increases, LLM performance improves.
Table~\ref{Tab:LLM_change} shows the facet generation performance of \textit{FD+M+E} when using different LLMs(7B, 13B, 30B).
We observe that even when utilizing LLM(7B, 13B), for editing instead of LLM(30B), there is still a notable performance improvement. 
Smaller LLMs tend to have higher performance in Term Overlap.
Therefore, LLM editing is an effective prompt technique regardless of the size of LLM.

In the case of \textit{E(few)}, which does not use the small model, there is a significant difference in performance between LLM(30B) and LLM(13B,7B).
That is, in few-shot inference, similar to the LLM benchmark, larger LLMs generally outperform smaller LLMs.
However, when combined with a small model, the difference between \textit{FD+M+E} is reduced, which shows that the role of the small model is crucial in generating facets of LLM.
The small model serves as an intermediary bridge between the query and facets because it has learned the distribution of facets through the training dataset. 
With the assistance of this small model, LLM can generate the desired facets.
We attempted various other LLMs, but it was challenging to find an LLM that excelled in all four metrics. 
This is considered a trade-off related to LLM size, and further research is needed to achieve better results in all metrics.


\end{document}